# Group Sparse Additive Models

Junming Yin                                              JUNMINGY@CS.CMU.EDU
Xi Chen                                                     XICHEN@CS.CMU.EDU
Eric P. Xing                                                EPXING@CS.CMU.EDU
School of Computer Science, Carnegie Mellon University, Pittsburgh, PA 15213, USA

## Abstract

We consider the problem of sparse variable selection in nonparametric additive models, with the prior knowledge of the structure among the covariates to encourage those variables within a group to be selected jointly. Previous works either study the group sparsity in the parametric setting (e.g., group lasso), or address the problem in the non-parametric setting without exploiting the structural information (e.g., sparse additive models). In this paper, we present a new method, called group sparse additive models (GroupSpAM), which can handle group sparsity in additive models. We generalize the $\ell_1/\ell_2$ norm to Hilbert spaces as the sparsity-inducing penalty in GroupSpAM. Moreover, we derive a novel thresholding condition for identifying the functional sparsity at the group level, and propose an efficient block coordinate descent algorithm for constructing the estimate. We demonstrate by simulation that GroupSpAM substantially outperforms the competing methods in terms of support recovery and prediction accuracy in additive models, and also conduct a comparative experiment on a real breast cancer dataset.

## 1. Introduction

The problem of sparse variable selection for high-dimensional data arises in a wide spectrum of domains, including signal processing, bioinformatics and computer vision. $\ell_1$-regularized methods, such as lasso (Tibshirani, 1996), are among the most widely used approaches for variable selection in linear models. Despite the popularity of linear models, a reliance on rigid parametric forms limits their ability to model nonlinear covariate effects. Additive models (Hastie & Tibshirani, 1990), where each additive component is a univariate smooth function of a single covariate, are nonparametric extensions of linear models and can offer a higher degree of flexibility. Variable selection in nonparametric additive models is more challenging as one needs to simultaneously select and fit component functions. While some recent progress has been made on this problem by proposing various functional penalties (Lin & Zhang, 2006; Bach, 2008; Ravikumar et al., 2009), none of these methods is capable of exploiting the structural information among the covariates that may exist as prior knowledge in many applications. The simplest case is a group structure where the covariates are partitioned into disjoint groups, and it is desirable to choose relevant covariates that are sparse at the group level. In the parametric setting, it has been shown that if such a group structure exists and is consistent with the true sparsity pattern of covariates, treating the whole group of covariates as a single unit in variable selection has the potential to increase the accuracy of the estimator (Huang & Zhang, 2010). However, to the best of our knowledge, there has been no successful attempts to investigate the benefit of considering the group sparsity in estimating nonparametric additive models. One example of such a situation occurs in biology, where the effects of multiple genes on the phenotype are nonlinear, and the goal is to identify a few genes from the same functional groups that are predictive of the phenotype.

In this paper, we fill this gap by presenting a new approach for variable selection in nonparametric additive models that can take advantage of the group structure *among the covariates*. The proposed method, called GroupSpAM, achieves *functional sparsity* at the level of groups by integrating the spirit of group lasso (Yuan & Lin, 2006) and the idea of sparse additive models (SpAM) (Ravikumar et al., 2009). Therefore, the GroupSpAM estimate combines the predictive power of SpAM that can model nonlinear covariate effects, with the advantage of group lasso that can incorpo-rate group structure to achieve better support recovery accuracy. Our empirical results provide convincing evidence of the expected benefits inherited from both threads of research. Although our main focus here is on regression, the framework can be easily extended to classification setting via generalized additive models.

In contrast to the ordinary group lasso, where there has been much confusion regarding orthonormality of covariates within a group, GroupSpAM is highly flexible in that no assumptions are made on the design matrices or on the covariance of component functions in each group. In this sense, GroupSpAM can be viewed as a nonparametric extension of the generalized group lasso (Friedman et al., 2010), which allows non-orthonormal covariates within a group, to additive models. However, while allowing nonlinear covariate effects and arbitrary covariance of component functions within a group gives rise to a new method that enjoys superior flexibility, there are new technical challenges in characterizing and computing the GroupSpAM estimate. To resolve these difficulties, we propose a novel optimization procedure to simultaneously conduct component selection and fitting at the level of groups. Specifically, our major contributions in this work include: (1) generalization of $\ell_1/\ell_2$ norm to $L_2$ function spaces as the sparsity-inducing penalty in GroupSpAM; (2) a necessary and sufficient condition for identifying functional sparsity at the group level (Theorem 2); (3) an efficient block coordinate descent algorithm tailored to handle the covariance of component functions within a group; (4) extension of GroupSpAM to the case of overlapping groups and a successful application to a real dataset.

## 2. Related Work

Lin & Zhang (2006) proposed the COSSO estimator, which uses the sum of the reproducing kernel Hilbert space (RKHS) norms of the component functions as the regularization penalty for simultaneous variable selection and model fitting in smoothing spline ANOVA models. Ravikumar et al. (2009) introduced a sparse version of additive models called SpAM by penalizing the sum of the $L_2(\mu)$ norms of the component functions. While these methods have been shown to be effective in estimating sparse nonparametric models, neither of them is able to take advantage of structural information. Liu et al. (2009a) extended SpAM to the multi-task setting; however, the group structure is imposed on the tasks instead of on the covariates being considered here.

Finally, it is important to emphasize that our work is substantially different from another line of research that also mixes group lasso and additive models (Bach, 2008; Meier et al., 2009; Huang et al., 2010). These approaches, in which each component function is expanded into a group of basis functions constructed from a *single* covariate, can only perform variable selection at an individual level. In contrast, in this paper, each group consists of a set of *several* covariates and our goal is to encourage those variables within a group to be selected jointly.

**Notation** Vectors and matrices are denoted by boldface letters, and estimates are denoted with a hat. For a random variable $X$ with distribution $\mu$ and a measurable function $f$ of $x$, $\|f\|$ denotes the $L_2(\mu)$ norm of $f$: $\|f\| = \sqrt{\mathbb{E}[f^2(X)]}$. With some slight abuse of notation we also use $\|\cdot\|$ to denote the $\ell_2$ norm of a real vector. For a set of random variables $X_1, \ldots, X_p$, let $\mathcal{H}_j = \{f_j \mid \mathbb{E}[f_j(X_j)] = 0, \|f_j\| < \infty\}, j = 1 \ldots, p$, with inner product on the space defined as $\langle f_j, g_j \rangle = \mathbb{E}[f_j(X_j)g_j(X_j)]$. Sometimes we also use $f_j := f_j(X_j)$ for simplicity. We use $\hat{\mathbf{f}}_j \in \mathbb{R}^n$ to represent the vector of evaluations of the fitted function $\hat{f}_j$ at the $n$ observed values of $X_j$, i.e., $\hat{\mathbf{f}}_j = [\hat{f}_j(x_{1j}), \ldots, \hat{f}_j(x_{nj})]^T$. A group of covariates is a subset $g \subset \{1, \ldots, p\}$ and a set of groups is denoted as $\mathcal{G}$. For any group $g$ and any vector $\mathbf{v}$, $\mathbf{v}_g = \{v_j\}_{j \in g}$. $\|\mathbf{f}_g\| = \sqrt{\sum_{j \in g} \|f_j\|^2}$. $d_g = |g|$ represents the cardinality of group $g$. We assume $\mathcal{G}$ is available in advance as prior knowledge and all covariates are covered by at least one group in $\mathcal{G}$.

## 3. Background

We begin by reviewing some important concepts on nonparametric regression and (sparse) additive models to set the stage for our method.

### 3.1. Smoothing for Nonparametric Regression

Let $X_1, \ldots, X_p$ be a set of random covariate variables and $Y$ be a real valued random response variable. Nonparametric regression is concerned with estimating the regression function

$$m(\boldsymbol{X}) = m(X_1, \ldots, X_p) = \mathbb{E}[Y \mid X_1, \ldots, X_p]$$

from a set of $n$ data samples $\{(\mathbf{x}^{(i)}, y^{(i)}) : \mathbf{x}^{(i)} \in \mathbb{R}^p, y^{(i)} \in \mathbb{R}, i = 1, \ldots, n\}$, without assuming any parametric form of $m(\boldsymbol{X})$, such as linearity in $X_1, \ldots, X_p$. We can write $Y = m(\boldsymbol{X}) + \epsilon$, where $\mathbb{E}[\epsilon] = 0$.

In the case of a single covariate ($p = 1$), $m(X) = \mathbb{E}[Y \mid X]$ is known as the orthogonal projection of $Y$ onto the linear space of all measurable functions of $X$ and can be written as

$$m(X) = PY, \qquad (1)$$

where $P$ is the conditional expectation operator $\mathbb{E}[\cdot \mid$

$X$]. By making the assumption that $m(x) = \mathbb{E}[Y \mid X = x]$ is a smooth function of $x$, we can estimate $m(x)$ using a class of smoothing estimators called the kernel smoothers

$$\widehat{m}(x) = \sum_{i=1}^{n} \ell_i(x) y^{(i)} = \ell(x)^T \mathbf{y}, \quad (2)$$

where $\ell_i(x) \propto K_h(|x^{(i)} - x|)$ and $K_h$ is a smoothing kernel function with the bandwidth $h$. Kernel smoothers are examples of linear smoothers because, for each $x$, the estimator $\widehat{m}(x)$ in (2) is a linear combination of $y^{(i)}$. Let $\hat{\mathbf{y}} \in \mathbb{R}^n$ be a vector of fitted values $\hat{y}^{(i)}$ at the observed $x^{(i)}$, one consequence of linear smoother is

$$\hat{\mathbf{y}} = \mathbf{S}\mathbf{y}, \quad (3)$$

where $\mathbf{S}$ is the so-called *smoother matrix* with $\mathbf{S}_{ij} = \ell_j(x^{(i)}), i, j = 1, \ldots, n$. A comparison of equations (1) and (3) reveals that the conditional expectation operator $P$ plays the role of smoother in the population setting, and consequently a natural estimate of $P$ is a linear smoother with smoother matrix $\mathbf{S}$. This one-dimensional smoother, called scatter smoother, is a building block for fitting more complicated models.

### 3.2. Additive Models

Although it is straightforward to generalize the one-dimensional smoothers to $p$-dimension case, it is well-known that smoothers break down in high dimensions due to the curse of dimensionality. This shortcoming motivates the study of additive models (Hastie & Tibshirani, 1990),

$$m(X_1, \ldots, X_p) = \alpha + \sum_{j=1}^{p} f_j(X_j), \quad (4)$$

where $f_1, \ldots, f_p$ are one-dimensional smooth component functions, one for each covariate. For simplicity and identification purposes, we assume $\alpha = 0$ and $f_j \in \mathcal{H}_j$ so that $\mathbb{E}[f_j(X_j)] = 0$ for each $j$. The optimization problem of additive models in the population setting is to minimize

$$L(\mathbf{f}) = \frac{1}{2}\mathbb{E}\left[\left(Y - \sum_{j=1}^{p} f_j(X_j)\right)^2\right], \quad (5)$$

over $\{\mathbf{f} : f_j \in \mathcal{H}_j\}$. The minimizers of (5) can be shown to satisfy

$$f_j = \mathbb{E}\left[(Y - \sum_{k \neq j} f_k) \mid X_j\right] := P_j\left(Y - \sum_{k \neq j} f_k\right), \quad (6)$$

where $P_j = \mathbb{E}[\cdot \mid X_j]$ is the projection operator onto $\mathcal{H}_j$. Replacing $P_j$ by a linear smoother with smoother matrix $\mathbf{S}_j$ in (6) immediately leads to a sample version of the above iterative procedure for fitting additive model:

$$\hat{\mathbf{f}}_j \leftarrow \mathbf{S}_j\left(\mathbf{y} - \sum_{k \neq j} \hat{\mathbf{f}}_k\right), j = 1, \ldots, p, 1, \ldots, p, \ldots \quad (7)$$

This simple algorithm is known as *backfitting* and is essentially a coordinate descent algorithm.

### 3.3. SpAM

However, additive models work well only in a low-dimensional $p \ll n$ setting. Ravikumar et al. (2009) proposed a new approach called SpAM for component selection in high-dimensional additive models. The idea is to impose a sparsity constraint on the index set of non-zero component functions via regularization:

$$\min_{\mathbf{f}: f_j \in \mathcal{H}_j} L(\mathbf{f}) + \lambda \Omega(\mathbf{f}), \quad (8)$$

where $\lambda > 0$ is the regularization parameter and $\Omega(\mathbf{f}) = \sum_{j=1}^{p} \|f_j\|$ behaves like an $\ell_1$ ball across different components to encourage functional sparsity. The stationary condition of (8) is given by

$$f_j = \left[1 - \frac{\lambda}{\|P_j R_j\|}\right]_+ P_j R_j, \quad (9)$$

where $R_j = Y - \sum_{k \neq j} f_k$ is the partial residual and $[\cdot]_+ = \max\{\cdot, 0\}$. A sample version of the algorithm can be obtained by inserting sample estimates into (9).

## 4. GroupSpAM

We are now equipped to present GroupSpAM. In this section, we assume $\mathcal{G}$ is a partition of $\{1, \cdots, p\}$, i.e., the groups in $\mathcal{G}$ do not overlap. The optimization problem of GroupSpAM in the population setting is formulated as

$$\min_{\mathbf{f}: f_j \in \mathcal{H}_j} L(\mathbf{f}) + \lambda \Omega_{\text{group}}(\mathbf{f}), \quad (10)$$

where $L(\mathbf{f})$ is the expected square error as in (5) and $\Omega_{\text{group}}(\mathbf{f})$ is the regularization functional penalty defined as

$$\Omega_{\text{group}}(\mathbf{f}) = \sum_{g \in \mathcal{G}} \sqrt{d_g} \|\mathbf{f}_g\| = \sum_{g \in \mathcal{G}} \sqrt{d_g} \sqrt{\sum_{j \in g} \mathbb{E}\left[f_j^2(X_j)\right]}.$$

The regularization term is the generalization of (scaled) $\ell_1/\ell_2$ penalty norm used in group lasso to $L_2$ function spaces. As in group lasso, this mixed norm induces sparsity at the level of groups: the whole group of functions $\mathbf{f}_g = \{f_j\}_{j \in g}$ are encouraged to be set to zero. If each group $g$ is a singleton, this formulation reduces to SpAM (Ravikumar et al., 2009); if each component function $f_j$ has a linear parametric form, the optimization problem in (10) is just the population setting of the ordinary group lasso formulation (Yuan & Lin, 2006).

To solve the optimization problem (10), a seemingly natural strategy is to employ a block coordinate descent algorithm: we optimize the objective functional in (10) with respect to a particular group of functions $\mathbf{f}_g$ at a time while all other groups of functions are kept fixed. However, as we allow arbitrary covariance of component functions within a group (see remark below for further discussion), there are two main obstacles for applying such algorithm to our problem: (1) it creates a new difficulty to characterize the thresholding condition for functional sparsity at the group level; (2) unlike the ordinary group lasso and SpAM, there is no longer a closed-form solution to the stationary condition for each group of functions $\mathbf{f}_g$, in the form of a soft-thresholding operator. Before describing the details of the algorithm, we first state the stationary condition that characterizes the optimum of $\mathbf{f}_g$.

**Theorem 1.** *Let $R_g = Y - \sum_{g' \neq g} \sum_{j' \in g'} f_{j'}(X_{j'})$ be the partial residual after removing all functions from the group $g$. The stationary condition of the problem (10) with respect to $\mathbf{f}_g$ while fixing all other groups $\{\mathbf{f}_{g'} : g' \neq g\}$ is*

$$f_j + \sum_{j' \in g : j' \neq j} P_j f_{j'} - P_j R_g + \lambda \sqrt{d_g} s_j = 0, \forall j \in g, \quad (11)$$

*where $\mathbf{s}_g = \{s_j \in \mathcal{H}_j\}_{j \in g}$ is a vector of functions belonging to the subgradient of $\|\mathbf{f}_g\|$:*

$$\mathbf{s}_g = \begin{cases} \{f_j / \|\mathbf{f}_g\|\}_{j \in g} & \text{if } \|\mathbf{f}_g\| \neq 0, \\ \{\mathbf{u}_g : \|\mathbf{u}_g\| \leq 1\} & \text{if } \|\mathbf{f}_g\| = 0. \end{cases}$$

*Proof.* The proof uses the calculus of variations in Hilbert space. See Appendix for details. □

Next, we prove a necessary and sufficient thresholding condition at the group level so that we can set a whole group of functions $\mathbf{f}_g$ to zeros if (12) holds.

**Theorem 2.** $f_j = 0 \; \forall j \in g$ *if and only if*

$$\sqrt{\sum_{j \in g} \mathbb{E}[(P_j R_g)^2]} \leq \lambda \sqrt{d_g}. \quad (12)$$

*Proof.* Necessity. If $f_j = 0 \; \forall j \in g$, then (11) reduces to $P_j R_g = \lambda \sqrt{d_g} s_j \; \forall j \in g$, with $\|\mathbf{s}_g\| \leq 1$. Thus

$$\sqrt{\sum_{j \in g} \mathbb{E}[(P_j R_g)^2]} = \lambda \sqrt{d_g} \sqrt{\sum_{j \in g} \mathbb{E}[s_j^2]} \leq \lambda \sqrt{d_g}.$$

Sufficiency. We prove by contradiction. If there exists an $f_j \neq 0, j \in g$ hence $\|\mathbf{f}_g\| \neq 0$, equation (11) becomes

$$P_j R_g = f_j + \sum_{j' \in g : j' \neq j} P_j f_{j'} + \frac{\lambda \sqrt{d_g}}{\|\mathbf{f}_g\|} f_j, \forall j \in g.$$

Without loss of generality, we assume $g = \{1, \cdots, |g|\}$. The set of equations in (11) can be succinctly written in the following equivalent form:

$$\mathbf{Q} R_g = \mathbf{J} \mathbf{f}_g + \frac{\lambda \sqrt{d_g}}{\|\mathbf{f}_g\|} \mathbf{f}_g, \quad (13)$$

where $\mathbf{Q}$ and $\mathbf{J}$ are a vector and matrix of conditional expectation operators, respectively, defined as

$$\mathbf{Q} = \begin{bmatrix} P_1 \\ P_2 \\ \vdots \\ P_{|g|} \end{bmatrix}, \mathbf{J} = \begin{bmatrix} I & P_1 & \cdots & P_1 \\ P_2 & I & \cdots & P_2 \\ \vdots & \vdots & \ddots & \vdots \\ P_{|g|} & P_{|g|} & \cdots & I \end{bmatrix}. \quad (14)$$

Hence

$$\sum_{j \in g} \mathbb{E}[(P_j R_g)^2] = \|\mathbf{Q} R_g\|^2 = \left\| \mathbf{J} \mathbf{f}_g + \frac{\lambda \sqrt{d_g}}{\|\mathbf{f}_g\|} \mathbf{f}_g \right\|^2$$

$$= \|\mathbf{J} \mathbf{f}_g\|^2 + \lambda^2 d_g + 2 \frac{\lambda \sqrt{d_g}}{\|\mathbf{f}_g\|} \langle \mathbf{J} \mathbf{f}_g, \mathbf{f}_g \rangle.$$

By the fact that

$$\langle \mathbf{J} \mathbf{f}_g, \mathbf{f}_g \rangle = \sum_{j \in g} \mathbb{E}[f_j^2 + f_j \sum_{j' \in g : j' \neq j} P_j f_{j'}]$$

$$= \sum_{j \in g} \mathbb{E}[f_j^2] + \sum_{j \in g} \sum_{j' \in g : j' \neq j} \mathbb{E}[f_j \mathbb{E}[f_{j'} \mid X_j]]$$

$$= \sum_{j \in g} \mathbb{E}[f_j^2] + \sum_{j \in g} \sum_{j' \in g : j' \neq j} \mathbb{E}[f_j f_{j'}]$$

$$= \mathbb{E}\left[ \left( \sum_{j \in g} f_j \right)^2 \right] \geq 0,$$

we conclude $\sum_{j \in g} \mathbb{E}[(P_j R_g)^2] \geq \lambda^2 d_g$. □

Following the standard additive models, we replace the conditional expectation operator $P_j$ by a linear smoother with smoother matrix $\mathbf{S}_j$ in the population conditions (12) and (13) to obtain a sample version of estimation procedure. Specifically, if the sample estimate of norm $\hat{\omega}_g$ in (15) is below the threshold $\lambda \sqrt{d_g}$, the whole group of functions are thresholded to zeros; otherwise, we estimate $\hat{\mathbf{f}}_g$ by solving the sample version of (13) in equation (16). See Algorithm 1 for details. The full block coordinate descent algorithm of the GroupSpAM procedure is described in Algorithm 2. Performing prediction on new data is essentially the same as in the standard additive models (Hastie & Tibshirani, 1990).

**Remark** As we don't restrict the covariance of component functions, for two distinct functions $f_j$ and $f_{j'}$

**Algorithm 1** Thresholding

1: **Input:** Partial residual $\widehat{\mathbf{R}}_g$, smoother matrices $\{\mathbf{S}_j : j \in g\}$, and tuning parameter $\lambda$.
2: **Output:** $\hat{\mathbf{f}}_g = \{\hat{\mathbf{f}}_j : j \in g\}$.
3: Estimate $P_j R_g$ by smoothing: $\widehat{\mathbf{P}}_j = \mathbf{S}_j \widehat{\mathbf{R}}_g, \forall j \in g$.
4: Estimate $\sqrt{\sum_{j \in g} \mathbb{E}[(P_j R_g)^2]}$ by

$$\widehat{\omega}_g = \sqrt{\frac{1}{n} \sum_{j \in g} \|\widehat{\mathbf{P}}_j\|^2}. \tag{15}$$

5: **if** $\widehat{\omega}_g \leq \lambda \sqrt{d_g}$ **then**
6:    Set $\hat{\mathbf{f}}_j = \mathbf{0}$, $\forall j \in g$.
7: **else**
8:    Estimate $\hat{\mathbf{f}}_g$ by solving the sample version of (13): iterate the following fixed point equation over $t$ until convergence

$$\hat{\mathbf{f}}_g^{(t+1)} = \left(\widehat{\mathbf{J}} + \frac{\lambda \sqrt{d_g}}{\|\hat{\mathbf{f}}_g^{(t)}\|/\sqrt{n}} \mathbf{I}\right)^{-1} \widehat{\mathbf{Q}} \widehat{\mathbf{R}}_g, \tag{16}$$

   where $\widehat{\mathbf{Q}}$ and $\widehat{\mathbf{J}}$ are matrices obtained by replacing each $P_j$ with $\mathbf{S}_j$ in $\mathbf{Q}$ and $\mathbf{J}$ (equation (14)), respectively.
9: **end if**
10: Center each $\hat{\mathbf{f}}_j$ by subtracting its mean

---

**Algorithm 2** Block Coordinate Descent

1: **Input:** Data $\mathbf{X} \in \mathbb{R}^{n \times p}, \mathbf{y} \in \mathbb{R}^n$, a partition $\mathcal{G}$ of $\{1, \ldots, p\}$, and tuning parameter $\lambda$.
2: **Output:** Fitted functions $\hat{\mathbf{f}} = \{\hat{\mathbf{f}}_j : j = 1, \ldots, p\}$.
3: Initialize $\hat{\mathbf{f}}_j = \mathbf{0}$ $\forall j$; pre-compute smoother matrices $\mathbf{S}_j$ $\forall j$.
4: Cycle though group $g \in \mathcal{G}$ until convergence:

  (1) Compute the partial residual $\widehat{\mathbf{R}}_g = \mathbf{y} - \sum_{g' \neq g} \sum_{j' \in g'} \hat{\mathbf{f}}_{j'}$

  (2) $\hat{\mathbf{f}}_g \longleftarrow \text{Thresholding}(\widehat{\mathbf{R}}_g, \{\mathbf{S}_j\}_{j \in g}, \lambda)$

---

in the same group, their covariance $\text{Cov}(f_j, f_{j'})$ is not forced to be zero. By the fact that

$$\text{Cov}(f_j, f_{j'}) = \mathbb{E}[f_j f_{j'}] = \mathbb{E}[f_j \mathbb{E}[f_{j'} \mid X_j]],$$

$P_j f_{j'} = \mathbb{E}[f_{j'} \mid X_j]$ is not restricted to be zero as well. Therefore, no closed-form solution to the stationary condition (11) is available and we solve its sample version by fixed point iteration in (16). On the other hand, if $P_j f_{j'}$ is assumed to be zero for *all* distinct $j, j' \in g$ in (11), the solution of $\hat{\mathbf{f}}_g$ can be directly obtained by applying a soft-thresholding operator: $\hat{\mathbf{f}}_j = \left[1 - \frac{\lambda \sqrt{d_g}}{\widehat{\omega}_g}\right]_+ \widehat{\mathbf{P}}_j$, $\forall j \in g$. However, such an assumption, though leading to a simple closed-form solution of $\hat{\mathbf{f}}_g$, is unrealistic in practice for within-group component functions.

## 5. GroupSpAM with Overlap

In this section, we allow overlap between the groups in $\mathcal{G}$ and we are interested in estimating an additive model whose support is a union of groups. We adopt the approach of Jacob et al. (2009) by introducing a set of latent functions $\mathbf{h}^g = \{h_j^g \in \mathcal{H}_j\}_{j \in g}$, one set for each group, and solve the following optimization problem

$$\min_{\mathbf{f}, \{\mathbf{h}^g\}_{g \in \mathcal{G}}} \quad L(\mathbf{f}) + \lambda \sum_{g \in \mathcal{G}} \sqrt{d_g} \|\mathbf{h}^g\|$$

$$\text{s.t.} \quad \sum_{g: j \in g} h_j^g = f_j, \, j = 1, \ldots, p. \tag{17}$$

The idea is to decompose each original component function to be the sum of a set of latent functions and then apply the functional group penalty to the decomposition. As a consequence, the covariates that are not in any selected group are removed and the resulting support is a union of groups. Once we eliminate $\mathbf{f}$ from (17), the problem reduces to

$$\min_{\{\mathbf{h}^g\}_{g \in \mathcal{G}}} \frac{1}{2} \mathbb{E}\left[\left(Y - \sum_{g \in \mathcal{G}} \sum_{j \in g} h_j^g(X_j)\right)^2\right] + \lambda \sum_{g \in \mathcal{G}} \sqrt{d_g} \|\mathbf{h}^g\|. \tag{18}$$

Algorithm 2 can be directly applied to solve (18) by treating each $\mathbf{h}^g$ as a block.

## 6. Experiments

### 6.1. Simulation Study

We generate covariates with compound symmetry covariance structure as follows: each covariate $X_j = (W_j + tU)/(1+t), j = 1, \ldots, p$, where $W_1, \ldots, W_p$ and $U$ are i.i.d from Uni(-2.5,2.5). For distinct covariates $X_j$ and $X_k$, $\text{corr}(X_j, X_k) = t^2/(1+t^2)$. The setting $t = 0$ corresponds to the case of independent covariates. The sample size $n = 150$, and the dimension of covariates $p = 200$ and 1000.

We then generate responses from an additive model in $\mathbb{R}^p$ with two groups of relevant component functions, each of size four (Table 1): $Y = m(\boldsymbol{X}) + \epsilon = \sum_{j=1}^{8} f_j(X_j) + \epsilon$, where $\epsilon \sim \mathcal{N}(0, \sigma^2)$. The component functions are drawn from Lin & Zhang (2006) and Ravikumar et al. (2009), and are appropriately scaled so that each group contains a function with relatively low variance. The standard deviation $\sigma$ of the noise is carefully chosen to give a signal-to-noise ratio $\sqrt{\text{Var}(m(\boldsymbol{X}))/\sigma^2} = 3$ in the case of uncorrelated covariates ($t = 0$). For each training data, we also generate equal-sized validation data and test data in the same manner. Validation datasets are used to choose the value of the parameter $\lambda$, and test datasets are used to measure the predication accuracy of the estimated models in terms of mean squared error (MSE).

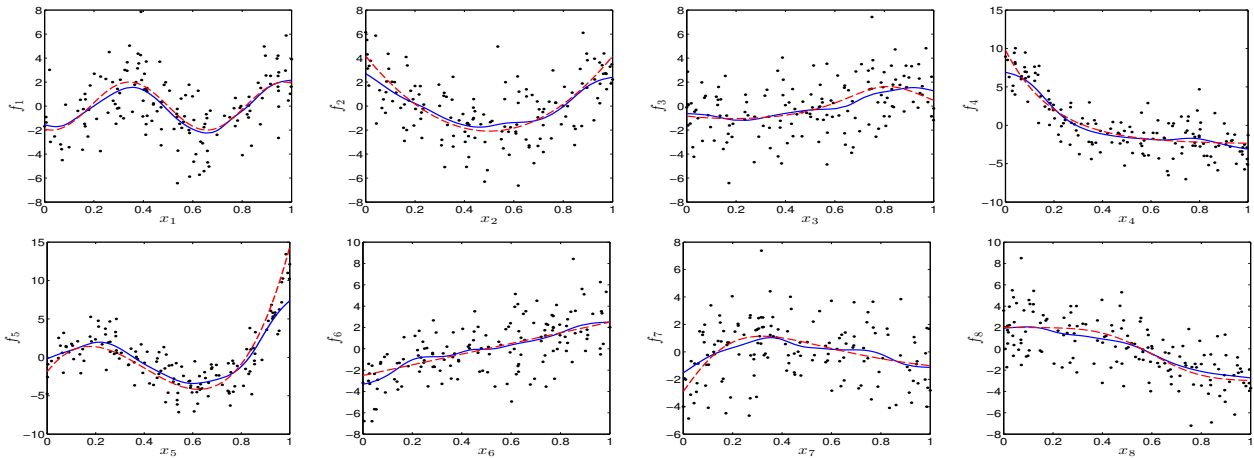

Figure 1. The estimated component functions (solid blue) and true component functions (dashed red) in one simulation with $p = 200, t = 0$. All the covariates are scaled to the interval $[0, 1]$. Black dots in each plot represent the partial residuals after removing the corresponding estimated component function.

| Component Functions | | | Variance |
|---|---|---|---|
| $f_1(x)$ | $=$ | $-2\sin(2x)$ | 2.10 |
| $f_2(x)$ | $=$ | $x^2$ | 3.47 |
| $f_3(x)$ | $=$ | $\frac{2\sin(x)}{2-\sin(x)}$ | 0.98 |
| $f_4(x)$ | $=$ | $\exp(-x)$ | 8.98 |
| $f_5(x)$ | $=$ | $x^3 + 1.5(x-1)^2$ | 14.57 |
| $f_6(x)$ | $=$ | $x$ | 2.08 |
| $f_7(x)$ | $=$ | $3\sin(\exp(-0.5x)$ | 0.80 |
| $f_8(x)$ | $=$ | $-5\phi(x, 0.5, 0.8^2)$ | 3.76 |

Table 1. The two groups of true component functions used in the simulation and their variances assuming $x$ is uniformly generated from $[-2.5, 2.5]$. $\phi(\cdot, \mu, \sigma^2)$ is the normal cumulative distribution function (cdf) with mean $\mu$ and standard deviation $\sigma$.

Table 2 summarizes the results of applying different methods to 100 simulated data for each value of $p = 200$, $p = 1000$ and $t = 0, 1, 2$. As expected, as the dimension ($p$) or the correlation ($t$) increases, the problem in general becomes more difficult. For all reported results, we assume a group structure with 50 or 250 blocks of 4 neighboring covariates for GroupSpAM and group lasso, and use Gaussian kernel smoothers with the plug-in bandwidths $h_j = 0.6\widehat{\text{sd}}(X_j)n^{-1/5}$ for $j$th covariate to implement GroupSpAM and SpAM. For group lasso, we use the implementation in the state-of-art package SLEP (Liu et al., 2009b); for COSSO[1], we use the Matlab code available from http://www4.stat.ncsu.edu/~hzhang/cosso.html. In almost all cases, GroupSpAM is able to recover all the true supports of covariates with higher precisions and have lower MSE on test data. Compared to the group lasso, GroupSpAM has significantly improved the predictive power of estimates, and also leads to much better support recovery when $t$ increases. The improvement is mainly in terms of variable selection as compared to SpAM, showing the benefit of considering the group sparsity in estimating additive models. It is particularly interesting to notice that the supports of component functions with relatively low variances, i.e., $f_3$ and $f_7$, are hard to be recovered as individual units by SpAM, but are much easier for GroupSpAM to select when combined with other components in the same group. Figure 1 shows the estimated component functions by GroupSpAM (solid blue line) versus the true component functions (dashed red line) in one typical simulation with $p = 200, t = 0$.

### 6.2. Breast Cancer Data

In our second experiment, we apply GroupSpAM with overlap to a breast cancer dataset (van de Vijver et al., 2002) to demonstrate the potential advantage of using additive models with group sparsity in a real-world problem. The dataset consists of gene expression measurements for 8,141 genes collected from 295 breast-cancer tumors (78 metastatic and 217 non-metastatic). We are interested in finding a sparse set of genes that can discriminate the two types of tumors. Instead of considering individual genes independently, a more powerful way is to build a predictive model that takes into account their pathway information. Some genes in a biological pathway are known to perform the same functionality in the cell, hence are more likely to be involved in the studied phenomenon in a group manner. Furthermore, each gene can participate in multiple pathways, and the studied phenomenon may possibly depend on the behavior of genes in a complex way. The GroupSpAM with overlap provides us with a natural and flexible way to incorporate these prior information into the biological analysis, where each group consists of the set of genes in a pathway and groups are potentially overlapping.

---
[1]For $p = 1000$, we get many warning messages in running COSSO, which suggests that COSSO might not scale well to high-dimensional settings (see also Table 2).

| $p$ | $t$ | method | precision | recall | size | $\#f_1$ | $\#f_2$ | $\#f_3$ | $\#f_4$ | $\#f_5$ | $\#f_6$ | $\#f_7$ | $\#f_8$ | MSE |
|---|---|---|---|---|---|---|---|---|---|---|---|---|---|---|
| 200 | 0 | GroupSpAM | 1.00 (0.00) | 1.00 (0.00) | 8.00 (0.00) | 100 | 100 | 100 | 100 | 100 | 100 | 100 | 100 | 7.22 (1.17) |
| | | SpAM | 0.85 (0.16) | 0.82 (0.11) | 8.17 (2.36) | 83 | 100 | 56 | 100 | 100 | 94 | 27 | 100 | 9.61 (2.22) |
| | | COSSO | 0.66 (0.27) | 0.42 (0.15) | 7.04 (5.48) | 6 | 1 | 27 | 100 | 50 | 61 | 3 | 88 | 28.29 (4.39) |
| | | GroupLasso | 0.95 (0.13) | 0.99 (0.05) | 8.64 (1.76) | 100 | 100 | 100 | 100 | 99 | 99 | 99 | 99 | 28.34 (3.10) |
| 200 | 1 | GroupSpAM | 0.96 (0.12) | 1.00 (0.00) | 8.52 (1.66) | 100 | 100 | 100 | 100 | 100 | 100 | 100 | 100 | 7.01 (1.29) |
| | | SpAM | 0.81 (0.20) | 0.60 (0.11) | 7.10 (6.26) | 94 | 82 | 4 | 98 | 100 | 3 | 10 | 86 | 9.10 (1.74) |
| | | COSSO | 0.33 (0.21) | 0.48 (0.19) | 16.08 (10.60) | 23 | 36 | 42 | 93 | 73 | 27 | 15 | 74 | 16.56 (2.93) |
| | | GroupLasso | 0.67 (0.38) | 0.65 (0.37) | 7.96 (6.39) | 64 | 64 | 64 | 64 | 65 | 65 | 65 | 65 | 20.69 (2.93) |
| 200 | 2 | GroupSpAM | 0.89 (0.19) | 0.99 (0.07) | 9.68 (4.20) | 100 | 100 | 100 | 100 | 98 | 98 | 98 | 98 | 7.26 (1.78) |
| | | SpAM | 0.71 (0.22) | 0.46 (0.12) | 5.90 (3.05) | 88 | 75 | 0 | 83 | 100 | 0 | 4 | 15 | 8.48 (1.74) |
| | | COSSO | 0.23 (0.12) | 0.41 (0.15) | 16.57 (7.47) | 11 | 61 | 22 | 90 | 76 | 10 | 10 | 47 | 13.72 (2.60) |
| | | GroupLasso | 0.13 (0.29) | 0.12 (0.24) | 5.60 (5.46) | 14 | 14 | 14 | 14 | 11 | 11 | 11 | 11 | 26.19 (3.11) |
| 1000 | 0 | GroupSpAM | 1.00 (0.00) | 1.00 (0.00) | 8.00 (0.00) | 100 | 100 | 100 | 100 | 100 | 100 | 100 | 100 | 7.21 (1.12) |
| | | SpAM | 0.86 (0.15) | 0.68 (0.14) | 6.60 (2.24) | 49 | 91 | 25 | 100 | 100 | 71 | 7 | 97 | 11.66 (2.73) |
| | | COSSO | 0.01 (0.00) | 0.97 (0.06) | 800.05 (1.16) | 93 | 100 | 97 | 100 | 100 | 100 | 84 | 100 | 36.59 (4.86) |
| | | GroupLasso | 0.93 (0.15) | 0.97 (0.11) | 8.80 (2.71) | 98 | 98 | 98 | 98 | 97 | 97 | 97 | 97 | 29.49 (4.23) |
| 1000 | 1 | GroupSpAM | 0.92 (0.17) | 0.99 (0.05) | 9.12 (3.00) | 99 | 99 | 99 | 99 | 100 | 100 | 100 | 100 | 7.34 (1.78) |
| | | SpAM | 0.71 (0.25) | 0.52 (0.13) | 8.62 (19.38) | 77 | 64 | 1 | 92 | 100 | 1 | 4 | 73 | 10.34 (2.25) |
| | | COSSO | 0.00 (0.00) | 0.00 (0.00) | 0.00 (0.00) | 0 | 0 | 0 | 0 | 0 | 0 | 0 | 0 | 21.95 (2.53) |
| | | GroupLasso | 0.31 (0.35) | 0.41 (0.41) | 8.92 (8.92) | 36 | 36 | 36 | 36 | 46 | 46 | 46 | 46 | 21.55 (2.81) |
| 1000 | 2 | GroupSpAM | 0.75 (0.30) | 0.97 (0.11) | 14.64 (14.76) | 95 | 95 | 95 | 95 | 100 | 100 | 100 | 100 | 8.10 (2.70) |
| | | SpAM | 0.69 (0.29) | 0.34 (0.13) | 6.04 (8.37) | 59 | 43 | 0 | 65 | 100 | 0 | 1 | 3 | 9.69 (2.30) |
| | | COSSO | 0.00 (0.00) | 0.00 (0.00) | 0.00 (0.00) | 0 | 0 | 0 | 0 | 0 | 0 | 0 | 0 | 26.30 (2.68) |
| | | GroupLasso | 0.02 (0.12) | 0.03 (0.16) | 4.36 (4.60) | 4 | 4 | 4 | 4 | 2 | 2 | 2 | 2 | 25.86 (3.06) |

*Table 2.* Comparison of different methods on simulated data. Shown in 4th, 5th and 6th column are the mean and standard deviation (shown in parenthesis) of precisions, recalls and sizes of the estimated supports, respectively. The symbol $\#f_j$ denotes the number of times $j$th covariate appears in the estimated models. The last column shows the mean and standard deviation of the mean squared errors (MSE) of the estimated models on test data sets.

Among all 8,141 genes, we focus on 3,510 of them that belong to at least one canonical pathway in the Molecular Signatures Database (Subramanian et al., 2005). We further reduce the gene set to top 300 genes most correlated with the type of tumor by applying the sure independence screening (Fan & Lv, 2008). This step, which excludes the most irrelevant genes, is a common practice for analyzing microarray data.

Overall, we obtain 1,369 covariates (gene expression levels) in 432 groups after covariate duplication. Since the dataset is heavily unbalanced, we adopt a balanced loss function, where each positive (negative) sample is weighted by the proportion of negative (positive) samples. To get the classification label from the non-parametric regression analysis, we simply take the sign of the predicted responses. Table 3 shows the results of GroupSpAM, SpAM and group lasso with overlap (Jacob et al., 2009) based on the balanced loss function by a 3-fold cross validation[2]. As we can see from Table 3, compared to SpAM, it achieves similar balanced error rates but with less selected genes and pathways, which could lead to an easier interpretation for genetic functional analysis. As compared to group lasso, GroupSpAM has an improved balanced error rate ($P = 0.054$), suggesting that a better predictive model can be built by using the more flexible additive model class. The functional relationship of the identified genes and pathways to breast cancer merits further investigation.

| fold | method | BER | #genes | #pathways |
|---|---|---|---|---|
| 1 | GroupSpAM | **0.353** | 55 | **196** |
| | SpAM | 0.362 | 91 | 266 |
| | GroupLasso | 0.384 | **44** | 238 |
| 2 | GroupSpAM | 0.358 | **44** | **243** |
| | SpAM | **0.349** | 109 | 302 |
| | GroupLasso | 0.365 | 56 | 248 |
| 3 | GroupSpAM | **0.326** | 74 | 149 |
| | SpAM | 0.333 | 101 | 209 |
| | GroupLasso | 0.346 | 76 | **138** |

*Table 3.* Comparison of different methods on the breast cancer dataset. BER refers to the balanced error rate, which is defined as the average of the errors in each tumor type. #genes denotes the number of *distinct* selected genes. #pathways denotes the number of selected pathways.

---
[2]When running COSSO, we ran into the same problem as in the simulation. Hence we left the results of COSSO.

# 7. Conclusions

In this paper, we propose a novel method for variable selection in nonparametric additive models when there exists a potentially overlapping group structure among the covariates. An efficient optimization algorithm is developed and promising results are obtained on both simulated and real data. An interesting future direction is to design new functional penalties to incorporate more rich structures among the covariates (e.g., hierarchical tree structure). Another future work is to investigate the asymptotic properties of the method, such as model selection and prediction consistency.

# Acknowledgments

This work is supported by NIH 1R01GM087694, NIH 1R01GM093156, and a Ray and Stephanie Lane Research Fellowship to JY. We would like to thank Guillaume Obozinski for helpful discussion of MKL and the anonymous reviewers for their valuable comments.

# Appendix: Proof of Theorem 1

*Proof.* Writing $L(\mathbf{f})$ in (5) as a functional that depends on $\mathbf{f}_g$ only, we obtain

$$L(\mathbf{f}_g) = \frac{1}{2}\mathbb{E}\bigg[\big(R_g - \sum_{j \in g} f_j(X_j)\big)^2\bigg].$$

Consider a perturbation of $L(\mathbf{f}_g)$ along the direction $\boldsymbol{\eta}_g = \{\eta_j \in \mathcal{H}_j\}_{j \in g}$,

$$L(\mathbf{f}_g + \epsilon\boldsymbol{\eta}_g) = \frac{1}{2}\mathbb{E}\bigg[\big(R_g - \sum_{j \in g}(f_j(X_j) + \epsilon\eta_j(X_j))\big)^2\bigg].$$

The first order approximation of $L(\mathbf{f}_g + \epsilon\boldsymbol{\eta}_g) - L(\mathbf{f}_g)$ is

$$\epsilon \sum_{j \in g} \mathbb{E}\bigg[\eta_j(X_j)\big(\sum_{j' \in g} f_{j'}(X_{j'}) - R_g\big)\bigg]$$

$$= \epsilon \sum_{j \in g} \mathbb{E}\bigg[\eta_j(X_j)\mathbb{E}\big[\big(\sum_{j' \in g} f_{j'}(X_{j'}) - R_g\big) \mid X_j\big]\bigg]$$

$$= \epsilon \sum_{j \in g} \bigg\langle \eta_j(X_j), \mathbb{E}\bigg[\big(\sum_{j' \in g} f_{j'}(X_{j'}) - R_g\big) \mid X_j\bigg] \bigg\rangle.$$

In the second step, we use the iterated expectation rule to condition on $X_j$; in the last step, noting $\mathbb{E}\big[\big(\sum_{j' \in g} f_{j'}(X_{j'}) - R_g\big) \mid X_j\big] \in \mathcal{H}_j$, we express the expectation in the form of inner product in $\mathcal{H}_j$ and thus obtain the gradient of $L(\mathbf{f}_g)$ as

$$\nabla L(\mathbf{f}_g) = \bigg\{\mathbb{E}\bigg[\big(\sum_{j' \in g} f_{j'}(X_{j'}) - R_g\big) \mid X_j\bigg]\bigg\}_{j \in g}.$$

Denote $\Omega_{\text{group}}(\mathbf{f}_g) = \sqrt{d_g}\|\mathbf{f}_g\|$. The stationary condition of $\mathbf{f}_g$ for minimizing $L(\mathbf{f}_g) + \lambda\Omega_{\text{group}}(\mathbf{f}_g)$ is

$$\mathbb{E}\bigg[\big(\sum_{j' \in g} f_{j'}(X_{j'}) - R_g\big) \mid X_j\bigg] + \lambda\sqrt{d_g}s_j = 0, \forall j \in g$$

and can be rewritten in the form of conditional expectation operator $P_j = \mathbb{E}[\cdot \mid X_j]$, as in equation (11). □